\documentclass[letterpaper]{article} 
\usepackage{aaai23}  
\usepackage{times}  
\usepackage{helvet}  
\usepackage{courier}  
\usepackage[hyphens]{url}  
\usepackage{graphicx} 
\usepackage{natbib}  
\usepackage{caption} 
\usepackage{algorithm}
\usepackage{algorithmic}

\usepackage{newfloat}
\usepackage{listings}
\DeclareCaptionStyle{ruled}{labelfont=normalfont,labelsep=colon,strut=off} 
\floatstyle{ruled}
\newfloat{listing}{tb}{lst}{}
\floatname{listing}{Listing}
\nocopyright

\usepackage{caption}
\usepackage{subcaption}
\usepackage[utf8]{inputenc} 
\usepackage[T1]{fontenc}    
\usepackage{amsfonts}       
\usepackage{nicefrac}       
\usepackage{xcolor}         

\usepackage{tikz}

\usepackage{amsmath}
\usepackage{amssymb}
\usepackage{mathtools}
\usepackage{amsthm}

\DeclareMathOperator*{\argmax}{arg\,max}

\DeclareMathOperator{\EX}{\mathbb{E}}

\theoremstyle{plain}
\newtheorem{theorem}{Theorem}

\newtheorem{lemma}{Lemma}

\theoremstyle{definition}
\newtheorem{definition}{Definition}

\theoremstyle{remark}

\newcommand{\TP}{\textit{Temporal-Phasing}}
\newcommand{\RP}{\textit{Reward-Phasing}}

\title{Task Phasing: Automated Curriculum Learning from Demonstrations}
\author{
   Vaibhav Bajaj,\textsuperscript{\rm 1} Guni Sharon,\textsuperscript{\rm 1} Peter Stone\textsuperscript{\rm 2}
}
\affiliations{
    \textsuperscript{\rm 1}Texas A\&M University\\
    \textsuperscript{\rm 2}The University of Texas at Austin and Sony AI\\
    vaibhavbajaj@tamu.edu, guni@tamu.edu, pstone@cs.utexas.edu

}

\begin{document}

\maketitle

\begin{abstract}
Applying reinforcement learning (RL) to sparse reward domains is notoriously challenging due to insufficient guiding signals.
Common RL techniques for addressing such domains include (1) learning from demonstrations and (2) curriculum learning. While these two approaches have been studied in detail, they have rarely been considered together. This paper aims to do so by introducing a principled task phasing approach that uses demonstrations to automatically generate a curriculum sequence. Using inverse RL from (suboptimal) demonstrations we define a simple initial task. Our task phasing approach then provides a framework to gradually increase the complexity of the task all the way to the target task, while retuning the RL agent in each phasing iteration. Two approaches for phasing are considered: (1) gradually increasing the proportion of time steps an RL agent is in control, and (2) phasing out a guiding informative reward function. We present conditions that guarantee the convergence of these approaches to an optimal policy. Experimental results on 3 sparse reward domains demonstrate that our task phasing approaches outperform state-of-the-art approaches with respect to asymptotic performance. 

\end{abstract}

\section{Introduction}

In domains with sparse reward signals, a reinforcement learning (RL) agent~\cite{portelas2020automatic,JMLR:v21:20-212} receives little to no signal regarding its performance. This phenomenon results in a limited ability to train and learn an optimized policy.
As a result, a stream of publications~\cite{nair2018overcoming,burda2018exploration,salimans2018learning,reddy2019sqil,Ecoffet2019GoExploreAN,zhu2020learning} presented various solutions towards effective RL in sparse reward settings. Two of the most common approaches considered in such cases are curriculum learning (CL)~\cite{cur_learning,Soviany2021CurriculumLA,cl_patho,JMLR:v21:20-212} and learning from demonstrations~\cite{salimans2018learning,zhu2020learning,nair2018overcoming}. In this paper, we investigate the impact of combining these two general approaches toward a CL continuum which is defined through demonstrations. We suggest applying inverse RL (IRL) to the provided demonstrations in order to obtain a dense reward function and/or a demonstration policy. The IRL outcomes are then used to define an initial simple task for a curriculum along with a continuous curriculum continuum. This continuum is defined by a convex combination between the initial and target tasks.
By training an RL agent on the resulting CL continuum (with progressively increasing complexity), we show---both theoretically and empirically---that an RL agent can be effectively trained to solve tasks that are challenging otherwise. We provide theoretical guarantees that the proposed CL will return an optimal policy under the assumption that the optimal policy as a function of the task continuum is continuous. 

Two domain-independent approaches are presented for defining the task continuum, namely \TP\ and \RP. \TP\ is designed to provide gradually increased control to the RL agent in lieu of a demonstrator agent obtained using IRL. This approach is shown to be especially effective in domains with no/few catastrophic actions, i.e., they provide options for recovering from actions that hinder performance. \RP, on the other hand, results in a task continuum where each task is some convex combination of an informative dense reward (provided by the IRL agent) and the target (sparse) reward. This approach is shown to be especially effective in domains where a meaningful guiding reward function can be extracted from the provided demonstrations. 
The theory provided in this paper proves that, under reasonable assumptions, the curriculum defined by \RP\ produces a monotonically non-decreasing policy return in expectation---a desirable outcome in many real-world environments. Such theory is a novel addition to existing CL theory, that otherwise focuses on the effects of different curriculum strategies on the convergence rate of the policy~\cite{weinshall2018curriculum,weinshall2020theory,yengera2021curriculum}.

Experimental results are provided for 3 continuous sparse reward domains. The results suggest that our proposed approaches are successful in converging to an optimized policy, whereas baseline RL algorithms fail to do so. Moreover, our proposed approaches also outperform prior approaches that apply CL, learning from demonstrations, or both.

In summary, the contributions of this paper are:
\begin{enumerate}
    \item Define two domain-independent \textit{Task Phasing} approaches that utilize sub-optimal demonstrations.
    \item Present state-of-the-art asymptotic performance for sparse reward RL domains.
    \item Present convergence guarantees for \textit{Task Phasing}, under reasonable assumptions.
    \item Prove that the curriculum defined by \RP\ produces a monotonically non-decreasing policy return in expectation. 
\end{enumerate}

\section{Preliminaries} 

In reinforcement learning (RL) an agent is assumed to learn through interactions with an underlying \textit{Markov decision process} (MDP) defined by: $\mathcal{S}$ -- the state space, $\mathcal{A}$ -- the action space, $\mathcal{P}(s_t,a,s_{t+1})$ -- the transition function of the form $\mathcal{P}:\mathcal{S} \times \mathcal{A} \times \mathcal{S} \mapsto [0,1]$, $R(s,a)$ -- the reward function of the form $R: \mathcal{S} \times \mathcal{A} \mapsto \mathbb{R}$, and $\gamma$ -- the discount factor. The agent is assumed to follow an internal policy, $\pi$, which maps states to actions, i.e., $\pi: \mathcal{S} \mapsto \mathcal{A}$.\footnote{Policies can also be defined as stochastic (soft policy), i.e., mapping states to a distribution over actions.} The agent's chosen action ($a_t$) at the current state ($s_t$) affects the environment such that a new state emerges ($s_{t+1}$) as well as some reward ($r_t$) representing the immediate utility gained from performing action $a_t$ at state $s_t$, given by $R(s_t, a_t)$. We use $\tau$ to denote a finite horizon trajectory of the form $\{s_0, a_0, r_0, s_1, ..., a_{t-1}, r_{t-1}, s_t\}$. 

In this paper, the MDP comprises of two parts, (a) the \textit{environment} and (b) the \textit{task}. 
The environment defines the state space ($\mathcal{S}$), action space ($\mathcal{A}$), transition function ($\mathcal{P}$),  and discount factor $\gamma$. The task defines the time steps when the RL agent is given control (at such time steps $a_t$ is determined by the RL agent). The task also defines the reward function $R$. Using this terminology allows sharing an environment across MDPs, each corresponding to a different task.

The expected sum of discounted rewards for a given task, $\mathcal{K}$, is denoted by $J^\mathcal{K}_\pi = \EX_{\tau \sim \pi} \sum_t \gamma^t R^\mathcal{K}(s_t,a_t)$ where $R^\mathcal{K}$ is the task reward function. The observed task rewards are used to tune a policy such that $J^\mathcal{K}_\pi$, is maximized. The policy $argmax_\pi[J^\mathcal{K}_\pi]$ is the optimal policy and is denoted by $\pi_\mathcal{K}^*$.\footnote{For some tasks $argmax_\pi[J^\mathcal{K}_\pi]$ is not unique. In such cases, $\pi_\mathcal{K}^*$ may refer to any optimal policy.}

In \textit{sparse reward domains} a relatively high proportion of the reward signals ($r_t$) are similar, making it challenging to obtain a meaningful gradient in $J^\mathcal{K}_\pi$ with respect to $\pi$. Such domains are notoriously challenging to solve (i.e., identify an optimal policy for)~\cite{nair2018overcoming}.

\subsection{Related work}

Prior RL approaches for solving sparse reward domains can be divided into three broad classes. (1) boosted exploration, (2) demonstration guidance, and (3) curriculum learning.

\subsubsection{Boosted exploration approaches}
Boosted exploration approaches \cite{Ecoffet2019GoExploreAN,burda2018exploration,poter,nair2018overcoming,NeurIPS21-ishand} are mostly domain-independent approaches to finding the optimal policy for a sparse reward task, by improving the manner in which exploration is conducted in the target environment. These methods use intrinsic motivation~\cite{barto2013intrinsic,oudeyer2009intrinsic,intrinsic} where the agent presents itself with exploration rewards that are different from those of the given task-specific rewards. Despite their effectiveness, the time to convergence for such approaches can be significant, as they require significant exploration. 

\subsubsection{Demonstration-guided approaches}
This class of algorithms~\cite{NIPS2016_cc7e2b87,Torabi2018BehavioralCF,fu2017learning,reddy2019sqil} attempt to learn policies that minimize the mismatch between the RL state-action visitation distribution and a demonstrator's state-action visitation distribution. The performance of the policies learned by such techniques is often limited by the demonstrator's performance, as they aim to only mimic the demonstrator and do not attempt to explore for better policies. On the other hand, Self-Adaptive Imitation Learning (SAIL)~\cite{zhu2020learning} proposed an off-policy imitation learning approach that can surpass the demonstrator by encouraging exploration along with distribution matching and replacing sub-optimal demonstrations with superior self-generated trajectories. Consequently, we consider this approach for comparison in our experiments. 

\subsubsection{Curriculum learning}
This class of algorithms~\cite{cur_learning,Soviany2021CurriculumLA,JMLR:v21:20-212} aims to break down complex tasks into simpler tasks. CL approaches often require a human domain expert to divide the task into simpler tasks and then design a curriculum that decides the sequence in which those tasks are learnt~\cite{cl_visual,cl_speech,cl_multi,cl_medical}. Recent work on CL~\cite{portelas2020automatic} aims to automate the curriculum design, but still requires some domain knowledge provided by a human expert, such as defining sub-tasks~\cite{pmlr-v100-portelas20a,matiisen2019teacher}, 
defining sub-goal conditions~\cite{poter,nair2018overcoming}, 
or scaling domain design features~\cite{Dennis2020EmergentCA}.  

\textbf{Note:}  \textit{Transfer learning} in RL~\cite{taylor2009transfer} is a subprocess of CL. As such, it is related to but not directly comparable to CL. For further details refer \citet{Navrekar}. 


\section{Task phasing} \label{Env_phasing}

Consider a sparse reward RL task defined by $\mathcal{K}$ and a given environment. Assume that $\mathcal{K}$ cannot be solved efficiently using common RL approaches~\cite{SAC,PPO} due to reward signal sparsity.
\textit{Task Phasing} addresses this inefficiency by introducing a set of simplified tasks with progressively increasing complexity (similar to CL). In contrast to most past CL algorithms, it does not rely on expert knowledge and/or domain-specific assumptions for defining the simplified tasks but defines them through demonstrations in a principled domain-independent way.

Assume some initial simplified task denoted $\mathcal{K}_s$, that can be efficiently solved by common RL algorithms, and a target (complex) task, $\mathcal{K}_f$, that cannot be solved by common RL algorithms. 
Next, consider a convex combination function $Con(\beta,\mathcal{K}_s,\mathcal{K}_f) = \beta \mathcal{K}_s + (1-\beta) \mathcal{K}_f$ which provides a task continuum, $\mathcal{K}_\beta$, with $\beta \in [0,1]$, between two given tasks $\mathcal{K}_s,\mathcal{K}_f$. 
We can now define the general task phasing curriculum procedure shown in Algorithm~\ref{alg:phasing}.

\begin{algorithm}[tb]
\caption{Task phasing curriculum learning.}
\label{alg:phasing}
\textbf{Input}: initial (simplified) task, $\mathcal{K}_s$; target (complex) task, $\mathcal{K}_f$; step size, $\alpha$; initial policy, $\pi$ \\
\textbf{Output}: optimized policy for $\mathcal{K}_f$, $\pi^*_f$
\begin{algorithmic}[1] 

\STATE $\pi^* \gets \text{train}(\pi, \mathcal{K}_s)$ \label{ln:train}
\STATE $\mathcal{K_\beta} \gets \mathcal{K}_s$
\STATE $\beta \gets 0$ 

\WHILE{$\mathcal{K_\beta} \ne \mathcal{K}_f$}
\STATE $\beta \gets \beta + \alpha$ 
 \STATE $\mathcal{K_\beta} \gets Con(\beta,\mathcal{K}_s,\mathcal{K}_f)$ 
 \STATE $\pi^* \gets \text{re-train}(\pi^*, \mathcal{K}_\beta)$ \label{ln:retrain}
\ENDWHILE
\STATE return $\pi^*$

\end{algorithmic}
\end{algorithm}


The `train' (Line~\ref{ln:train}) and `re-train' (Line~\ref{ln:retrain}) functions can be implemented with any off-the-shelf RL algorithm (assuming sufficient exploration, see Sec~\ref{sec:theory} for details). Note that this general approach only introduces a single hyper-parameter over the underlying RL solver, the $\alpha$ step size. A large value of $\alpha$ is desirable as it quickly leads to the optimization of the RL policy on the target task, $\mathcal{K}_f$. But, this may lead to training instability due to a large shift in the objective between consecutive tasks obtained from the task continuum. Determining the dynamic upper bound on the value of $\alpha$, that maintains training stability, at any instance of the \textit{Task Phasing} process is an interesting problem that we leave to future work. We found (empirically) that using a larger value of $\alpha$ during the initial portion of phasing and a smaller value as $\beta \rightarrow 1$ results in stable training.

Algorithm~\ref{alg:phasing} raises two questions that need to be addressed. 
\begin{enumerate}
    \item How can we obtain the initial simplified task ($\mathcal{K}_s$) for a general (domain-independent) MDP?
    
    \item How can we define the task continuum, i.e., $Con(\beta,\mathcal{K}_s,\mathcal{K}_f)$ for any $\beta \in [0,1]$?
\end{enumerate}

We address these questions by assuming a set of trajectories, $\mathcal{D}$, obtained from a (sub-optimal) demonstrating policy. 

\subsection{Temporal Task Phasing} \label{sec:temporal}

The \TP\ approach can be viewed as gradually shifting the control from a demonstrating policy to a learned RL policy. This approach assumes that a demonstrator policy, $\pi^d$, can be retrieved from $\mathcal{D}$, e.g., by using \textit{imitation learning}~\cite{NIPS2016_cc7e2b87,fu2017learning}. 

\TP\ follows some internal logic for determining in which time steps the RL-agent is given control, instead of the demonstrator, during online exploration. A similar approach was previously proposed~\cite{sheel} for smooth policy transitions. At any state, if the RL-agent is in control, the chosen action follows the RL internal policy. If, by contrast, the demonstrator is given control, then $\pi^d$ is followed. Transitions that follow the demonstrator's policy might also be used to train the RL-agent if off-policy learning~\cite{pmlr-v28-levine13} is enabled.

\paragraph{Initial simplified task ($\mathcal{K}_s$).}
$\mathcal{K}_s$ is defined by setting the probability of providing control to the RL-agent at any state to be 0.

\paragraph{task continuum ($Con(\beta,\mathcal{K}_s,\mathcal{K}_f)$).}
The environment continuum is defined by setting the probability of providing control to the RL-agent at any state to be $\beta$. Relevant protocols include:

\hspace{5mm}\textbf{V1: random step}
\[\forall t,~ \pi_{t} = \begin{cases} \mbox{$\pi^{RL}$,} & \mbox{$U[0,1) < \beta$} \\ \mbox{$\pi^{d}$,} & \mbox{else} \end{cases}\]
where $U[0,1)$ is a random value drawn from a uniform distribution in the range $[0,1)$.

\hspace{5mm}\textbf{V2: random $m$ steps}
\[\forall t:(t~\text{mod}~m)=0,~ \pi_{t:t+m-1} = \begin{cases} \mbox{$\pi^{RL}$,} & \mbox{$U[0,1) < \beta$} \\ \mbox{$\pi^{d}$,} & \mbox{else} \end{cases}\]

\hspace{5mm}\textbf{V3: fixed steps}
\[\forall t,~ \pi_{t} = \begin{cases} \mbox{$\pi^{d}$,} & \mbox{$(t~\text{mod}~\beta T)=0$} \\ \mbox{$\pi^{RL}$,} & \mbox{else} \end{cases}\]
where $T$ is the total number of steps for the episode. 

We implemented and experimented with all three variants and found `V1: random step' to be superior in most cases. Refer to Fig. A.2 in the Appendix for a comparison of the performance of the \TP\ variants.

\paragraph{Limitations:} \TP\ is expected to perform poorly in domains where actions have unrecoverable outcomes. For example, consider a robotic gripper arm task where the goal is to carry an object from one location to another. The demonstrator's policy never opens the gripper's palm, steadily holding the object. An exploratory RL agent, by contrast, will try various actions (including opening the gripper's palm) that result in unrecoverable situations and reduced probability of reaching the goal state. This probability diminishes exponentially as the exploratory RL agent is provided control in more time steps (factoring in the probabilities of avoiding unrecoverable actions per step). This phenomenon is especially harmful in sparse reward domains as a guiding signal from the goal state is rarely observed. 

\subsection{Reward Task phasing} \label{sec:reward}

The \RP\ approach can be viewed as gradually shifting the Reward function from a dense, informative reward signal to the true (sparse) reward. The dense reward is initially used to guide the RL agent such that it learns to imitate the demonstrator. Next, the dense reward is gradually phased out leaving the true reward as the only policy guiding signal. Doing so allows the RL agent to learn optimized policies that may diverge from the demonstrator. 
This approach assumes that an initial dense reward function, $R^{d}$, can be retrieved from $\mathcal{D}$, e.g., by using \textit{inverse RL}~\cite{fu2017learning,abbeel2004apprenticeship}. It further assumes that $R^{d}$ and the target reward function are of similar scale. This, however, is not a limiting assumption as $R^{d}$ can be scaled arbitrarily with no impact on the IRL efficiency.

\paragraph{Initial simplified task ($\mathcal{K}_s$).}
Let $R^{d}$ be the learnt dense reward function and let $R^f$ be the target (sparse) reward function.
$\mathcal{K}_s$ is defined by setting the reward function $R^s=R^{d} + R^f$.

\paragraph{Task continuum ($Con(\beta,\mathcal{K}_s,\mathcal{K}_f)$).}
The task continuum, $Con(\beta,\mathcal{K}_s,\mathcal{K}_f)$), is defined by a phased reward function that is a combination of $R^{d}$ and $R^f$. Relevant protocols include:

\hspace{5mm}\textbf{V1: constant phasing}
$$R^\beta= (1-\beta) R^{d} + R^f$$

\hspace{5mm}\textbf{V2: random phasing}
\[R^\beta = \begin{cases} \mbox{$R^{d}+R^f$,} & \mbox{$U[0,1) > \beta$} \\ \mbox{$R^f$,} & \mbox{else} \end{cases}\] 

We implemented and experimented with both variants and found `V1: constant phasing' to be superior in most cases. Refer to Fig. A.4 and Fig. A.5 in the Appendix for a comparison of the performance of the \RP\ variants.

\paragraph{Limitations:} \RP\ is expected to perform poorly in domains where the initial dense reward function directs the RL agent towards a locally optimal policy (with respect to the target reward). For example in a capture-the-flag game (e.g., our PyFlag domain as described in Appendix A.1), an initial reward that is biased towards guarding the player's flag will make it challenging to learn a policy that steals the enemy flag.
This is less of an issue for \TP\ as it trains a ``demonstrator'' surrogate function that can be queried for states that lay outside of the demonstration's state distribution, allowing the RL agent to explore further from the provided demonstrations.

\subsection{Convergence condition}
\label{sec:theory}

\begin{definition}[$RL_\epsilon$]\label{def:rl_e}
Define $RL_\epsilon$ to be an RL algorithm that explores and returns the optimal policy within some bounded $\epsilon$ KL-divergence from a given stochastic policy, $\pi$. That is, $RL_\epsilon$ will return  $$\pi^{b*}=\argmax_{\pi'|\EX_{s\sim \pi'}[KL(\pi(s),\pi'(s))] \le \epsilon} J_\pi$$ where $\pi^{b*}$ is the optimal policy within the exploration bound. 
\end{definition}
For simplicity of presentation, $KL(\pi,\pi')$ is used to represent $\EX_{s\sim \pi}[KL(\pi(s),\pi'(s))]$ hereafter. 

Examples of $RL_\epsilon$ include Trust Region Policy Optimization~\cite{trpo} and Proximal Policy Optimization~\cite{PPO}.

Consider two tasks $\mathcal{K}_1$ and $\mathcal{K}_2$ each with an affiliated optimal policy $\pi^*_1$ and $\pi^*_2$. 
It is easy to see that $RL_\epsilon$ will return
$\pi^*_2 \gets \text{re-train}(\pi^*_1, \mathcal{K}_2)$ if $KL(\pi^*_1, \pi^*_2)] \le \epsilon$.
This is because the optimal policy $\pi^*_2$ is within the exploration range from the initial policy $\pi^*_1$.

Consider applying Algorithm~\ref{alg:phasing} for a given $\mathcal{K}_s$, $\mathcal{K}_f$, and some RL algorithm, $RL_\epsilon$. Assume that $\pi^*_s$ is within the initial KL-divergence bound of $RL_\epsilon$, yet $\pi^*_f$ is not.
Further assume that, $KL(\pi^*_s, \pi^*_f) > \epsilon$. That is, $RL_\epsilon$ might fail to identify 
$\pi^*_f \gets \text{re-train}(\pi^*_s, \mathcal{K}_f)$.

\begin{lemma}[Convergence] \label{lem:convergence}
if the optimal policy, $\pi^*_\beta$, as a function of $\beta$ (the optimal policy for $\mathcal{K}_{\beta}$) is continuous for $\beta \in [0,1]$ and a given $\mathcal{K}_\beta = Con(\beta,\mathcal{K}_s, \mathcal{K}_f)$ function, then Algorithm~\ref{alg:phasing} with a small enough $\alpha$, using $RL_\epsilon$ as the underlying solver, will converge on $\pi^*_f$ within $1/\alpha$ iterations. 
\end{lemma}
\begin{proof}
If the optimal policy as a function of $\beta$ is continuous then there must exist a fine enough $\beta=\{\beta_0=0, \beta_1,...,\beta_n=1\}$ decomposition such that $\forall i \in \{0,...,n-1\}, KL(\pi^*_{\beta_i},\pi^*_{\beta_{i+1}}) \le \epsilon$. 
Following Definition~\ref{def:rl_e}, $RL_\epsilon$ will return $\pi^*_{\beta_{i}} \gets \text{re-train}(\pi^*_{\beta_{i-1}}, \mathcal{K}_{i})$ for every $i \in \{1,...,n\}$ where $\beta_{n} = 1$, i.e., $\mathcal{K}_{n}=\mathcal{K}_f$. That is, $RL_\epsilon$ will find all optimal policies along the resulting curriculum, until $\pi^*_f$.
\end{proof}

\begin{figure}[]
\centering
\includegraphics[width=0.7\columnwidth, height=2.5cm]{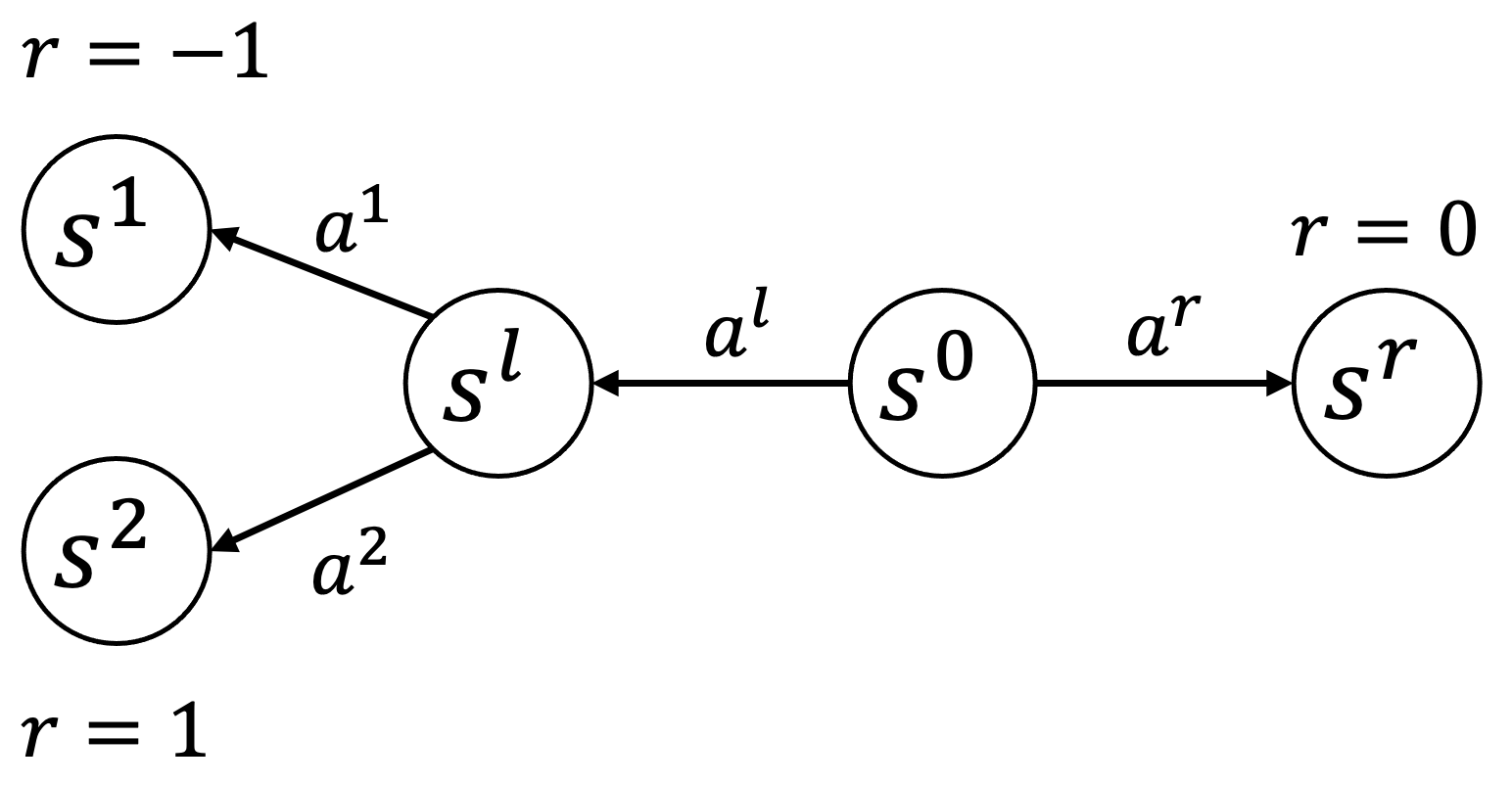}
\caption{An example MDP where both \TP\ and \RP\ using sub-optimal demonstrations result in a non-continuous policy space. The rewards shown in the MDP represent the target reward, $R^f$. \textbf{Temporal phasing}: assume $\pi^d([s^0,s^l])=[a^r,a^1]$, for $\beta<0.5$, $\pi^*_\beta(s^0)=a^r$ yet for $\beta>0.5$, $\pi^*_\beta(s^0)=a^l$. \textbf{Reward phasing}: assume $R^{d}([(s^0,a^r),(s^0,a^l),(s^l,a^1),(s^l,a^2)])=[2,0,0,0]$ and $\gamma = 1$. Using constant phasing, $R^\beta= (1-\beta) R^{d} + R^f$, for $\beta<0.5$, $\pi^*_\beta(s^0)=a^r$ yet for $\beta>0.5$, $\pi^*_\beta(s^0)=a^l$.}
\label{fig:not conv}
\end{figure}


A continuous $\pi^*_\beta$ function for both the \TP\ (Sec~\ref{sec:temporal}) and \RP\ (Sec~\ref{sec:reward}) approaches does not hold for the case of a sub-optimal demonstrator (from which $\mathcal{D}$ is sampled). Figure~\ref{fig:not conv} presents a counter examples for both spaces (Temporal and Reward phasing). 
As a result, in the worst case, an arbitrary small step in the task continuum can lead to a re-train procedure that is as hard as training a policy from scratch. 
Fortunately, this issue can be mitigated by considering a MaxEnt soft policy objective~\cite{SAC} which smoothens the policy continuum, $\pi^*_\beta$ resulting in a continuous shift between the policies in the re-train procedure. 
Reconsider the example from Figure~\ref{fig:not conv}, with the addition of an entropy maximizing term. That is, $\pi^*_\beta = \sum_t r_t + \mathcal{H}(\pi(\cdot|s_t))$. The reader is encouraged to validate that, in this case, the $\pi^*_\beta$ function is indeed continuous. 
Hence, for the reported experiments we choose an $RL_\epsilon$ algorithm that includes an entropy coefficient parameter (see Section~\ref{Sec:settings}).

\subsection{Theoretical results for \RP} 

Next we show that Algorithm~\ref{alg:phasing} results in monotonically non-decreasing policy performance (assuming $\pi^*_{\beta+\alpha} \gets \text{re-train}(\pi^*_\beta,\mathcal{K}_{\beta+\alpha})$ is found in each iteration). This result is important and useful, since assuming the initial performance is equivalent to that of the demonstrator, it guarantees that the learned policy never underperforms the demonstrator (in expectation). Note that this result relates to V1 (constant \RP). Nonetheless, the same result can be extended to apply for V2 following the fact that the affiliated reward functions are equal in expectation. That is,
$\EX[R^\beta_{V2}]=(1-\beta)(R^d+R^f)+\beta R^f=(1-\beta) R^{d} + R^f=R^\beta_{V1}$.

Define $\mathcal{R}_\pi^{d} := \EX_{\tau \sim \pi} \sum_t \gamma^t R^{d}(s_t, a_t)$.\\
Define $\mathcal{R}_\pi^{f} := \EX_{\tau \sim \pi} \sum_t \gamma^t R^{f}(s_t, a_t)$.\\
Consequently, we can rewrite 
$$\pi^*_\beta := \argmax_\pi (1-\beta) \mathcal{R}_\pi^{d} + \mathcal{R}_\pi^{f}$$

\begin{theorem}[Monotonic improvement] \label{theorem:theory_rew_phasing}
$\mathcal{R}_{\pi^*_\beta}^{f}$ is monotonically non-decreasing with $\beta$.  
\end{theorem}
\begin{proof}
by contradiction, assume some $\pi=\pi^*_\beta, ~\pi'=\pi^*_{\beta'}$, with $\beta < \beta'$, for which $\mathcal{R}_{\pi}^{f} > \mathcal{R}_{\pi'}^{f}$

\textbf{Case 1: $\mathcal{R}_{\pi}^{d} \ge \mathcal{R}_{\pi'}^{d}$}\\
This would imply that $(1-\beta') \mathcal{R}_{\pi'}^{d} + \mathcal{R}_{\pi'}^{f} < (1-\beta') \mathcal{R}_{\pi}^{d} + \mathcal{R}_{\pi}^{f}$ contradicting the assumption that $\pi'=\pi^*_{\beta'}$.

\textbf{Case 2: $\mathcal{R}_{\pi}^{d} < \mathcal{R}_{\pi'}^{d}$}\\
Let $\alpha = \beta' - \beta$. Then
\begin{equation}
   (1-\beta') \mathcal{R}_{\pi'}^{d} + \mathcal{R}_{\pi'}^{f} = (1-\beta) \mathcal{R}_{\pi'}^{d} + \mathcal{R}_{\pi'}^{f} -\alpha \mathcal{R}_{\pi'}^{d}
\end{equation}
\begin{equation}
   \le (1-\beta) \mathcal{R}_{\pi}^{d} + \mathcal{R}_{\pi}^{f} -\alpha \mathcal{R}_{\pi'}^{d}
\end{equation}
\begin{equation}
   < (1-\beta) \mathcal{R}_{\pi}^{d} + \mathcal{R}_{\pi}^{f} -\alpha \mathcal{R}_{\pi}^{d}=(1-\beta') \mathcal{R}_{\pi}^{d} + \mathcal{R}_{\pi}^{f}
\end{equation}
contradicting the assumption that $\pi'=\pi^*_{\beta'}$. \\
(Eq~2) because $\pi=\pi^*_\beta$ \quad (Eq~3) by the Case 2 assumption
\end{proof}

\section{Experiments and Results}   \label{sec:experiments}

The experiments are designed to study the performance of our two \textit{Task Phasing} variants when paired with a state-of-the-art RL solver. Specifically, they are designed to answer the following questions.
\begin{enumerate}
    \item Can our \textit{Task Phasing} variants learn an optimized policy in sparse-reward domains where the paired RL algorithm (without task phasing) cannot? \label{Q1}
    
    \item Can the proposed task phasing variants learn a policy that outperforms the demonstrator, where the demonstrator's policy is used for collecting the demonstrations in $\mathcal{D}$? \label{Q2}
    
    \item Are the limitation reported for each of our two \textit{Task Phasing} variants (Sections~\ref{sec:temporal} and~\ref{sec:reward}) observed empirically? \label{Q3}

    \item How does \textit{Task Phasing} compare to state-of-the-art algorithms designed to run in sparse reward domains and/or leverage demonstrations? \label{Q4}
\end{enumerate}

Our results provide a positive answer to Questions 1--3 and show clear advantages over previous state-of-the-art algorithms with respect to Question 4. 
In order to support full reproducibility of the reported results, our codebase along with detailed running instructions are provided online~\footnote{\url{https://github.com/ParanoidAndroid96/Task-Phasing.git}}.
The experiments are carried out in three continuous control, sparse reward environments, namely:
\begin{itemize}
    \item \textbf{PyFlags~\cite{pyflags} (GNU General Public License v3.0)} - This task requires a tank to fire at an opponent tank and capture its flag while defending its own flag.
    \item \textbf{FetchPickAndPlace-v1(P\&P)~\cite{Plappert2018MultiGoalRL}~(MIT license)} - This task requires a robotic arm to grab a randomly spawning block on a tabletop and carry it to a random goal location within the arms reach.
    \item \textbf{FetchSlide-v1(FS)~\cite{Plappert2018MultiGoalRL} (MIT license)} - This task requires a robotic arm to push a randomly spawning block on a tabletop such that it slides to a stop at a random goal location outside the robotic arms reach.
\end{itemize}

The FetchPickAndPlace and FetchSlide domains are used as sparse reward benchmark domains in the HER with demonstrations paper~\cite{nair2018overcoming}. All 3 domains are of special interest as they correspond to the limitations reported for our \textit{Task Phasing} variants. Specifically, the PyFlags domain (capture the flag) has many local optimums in the policy space, which is expected to have a stronger negative impact on \RP\ compared to \TP. On the other hand, the two Fetch domains have unrecoverable actions, which are expected to have a stronger negative impact on \TP\ compared to \RP. 

A snapshot from each of the reported domains is provided in Figure A.1. The technical description for the domains is provided in Appendix A.1. All experiments are repeated with 3 different random seeds and the mean of their results is reported along with a 1-$\sigma$ shaded region. 

\subsection{Settings} \label{Sec:settings}
We use a common $RL_\epsilon$ algorithm, denoted Proximal Policy Optimization (PPO)~\cite{PPO}, as our RL algorithm (for computing `train', Line~\ref{ln:train}, and `re-train', Line~\ref{ln:retrain}, in Algorithm~\ref{alg:phasing}). All algorithms used in the experiments apply the ADAM optimizer for training. The hyperparameters values for our approach as well as for the baseline algorithms are provided in Appendix A.2.

\subsubsection{Temporal phasing}
\TP\ requires an online demonstrator. The demonstrator can be learnt from demonstrations using techniques such as IRL~\cite{fu2017learning} or GAIL~\cite{NIPS2016_cc7e2b87}. In order to focus the study on the phasing approach, we skip the IL phase and directly use a suboptimal rules-based demonstrator (same one used to collect the demonstrations). We report results for the best performing \TP\ variant `V1: random step'. We provide a comparison of the various variants of \TP, as well as ablation studies conducted on them, in the Appendix A.3.
We found that using a dynamic $\alpha$ step size performed better compared to using a static one. In our dynamic setting, $\beta$ is incremented by $\beta \gets \beta+\alpha$ only if the average performance of the current policy ($\pi_\beta$) is greater than a set threshold. The threshold values for each domain are provided in Appendix A.2.
Also, for \TP\ we used importance sampling to train the RL-policy over transitions that follow from the demonstrator control (off-policy training). We used a PPO importance sampling approach that was introduced by~\cite{pmlr-v28-levine13}. However, instead of using Differential Dynamic Programming to generate ``guiding samples''~\cite{pmlr-v28-levine13}, we use the demonstrator's actions.

\subsubsection{\RP.}
\RP\ requires a dense reward function to provide initial guidance to the RL policy. In our approach, we use the Adversarial IRL~\cite{fu2017learning} approach to learn the dense reward function for the task. Results are reported for the best performing \RP\ variant (V1: constant phasing). Similar to \TP, we provide a comparison of the various variants of \RP\ in Appendix A.3. In order to allow reasonable running times, we employ a fixed interval approach, where $\beta \gets \beta + \alpha$ is applied every 200 training episodes. That is, Line~\ref{ln:retrain} in Alg~\ref{alg:phasing} trains for a fixed 200 episodes (not necessarily until convergence to $\pi_\beta^*$).

The experiments conducted indicate that the performance of both \TP\ and \RP\ is sensitive to the $\alpha$ parameter (step size), learning rate, and entropy coefficient. However, annealing these parameters throughout the learning process leads to stable (insensitive) results for both approaches. Further details on the hyperparameters used can be found in Appendix A.2.

\subsubsection{Baselines}
We compare the proposed approaches against the following baseline algorithms that are considered state-of-the-art for sparse reward domains:

\paragraph{Hindsight Experience Replay (HER) with demonstrations~\cite{nair2018overcoming}.}
This approach automatically generates a learning curriculum. It does so by setting intermediate goals in the environment that the policy can reach and providing rewards based on how close the agent is to a goal state. This approach requires some domain knowledge as the goal state for the environment must be defined. We find that the agent learns to retrieve the flag when the goal state for the PyFlag task is set such that the agent is positioned at its own base with the opponent's flag in its possession (distance to red flag = 0) and its own flag is not in the opponent's possession. When computing rewards, a weight of 1.0 is provided for possession of the opponent's flag and preventing the opponent from taking the agent's flag, in order to encourage offensive and defensive strategies. A lower weight, 0.1, is provided for the agent being positioned at its base so that it will prioritize attempting to steal the flag instead of staying close to its base. The goal state and the reward for the Fetch domains is the same as in~\cite{nair2018overcoming}, with the exception that the reward is scaled up from $\{-1,0\}$ to $\{0,1\}$.

\paragraph{Self-adaptive Imitation Learning (SAIL)~\cite{zhu2020learning}.} 
Instead of relying purely on exploration or on exploiting demonstrations, this approach aims to effectively strike a balance between exploiting sup-optimal demonstrations and efficiently exploring the environment to surpass the performance of the demonstrator. It maintains two replay buffers, for caching teacher demonstrations and self-generated transitions, respectively. During iterative training, SAIL dynamically adds high-quality self-generated trajectories into the teacher demonstration buffer.

\paragraph{Random Network Distillation (RND)~\cite{burda2018exploration}.}
This approach encourages the exploration of new states in the environment but does not rely on demonstrations. The exploration boost is provided by combining an intrinsic reward with an extrinsic (true) reward. The intrinsic reward is the error of a neural network predicting features of the observations given by a fixed randomly initialized neural network. 

\paragraph{Other relevant baselines} algorithms include: HER+DDPG \cite{andrychowicz2017hindsight}, BC+HER, GAIL~\cite{ho2016generative}, DAC~\cite{kostrikov2018discriminator}, DDPGfD~\cite{vecerik2017leveraging}, POfD~\cite{kang2018policy}. However, these baselines are omitted from our reported results as they were shown~\cite{nair2018overcoming,zhu2020learning,burda2018exploration} to be inferior to the aforementioned 3 state-of-the-art baselines.

\subsection{Results} \label{Sec:results}

\begin{figure*}[ht!]
    \centering
	\includegraphics[width=1.0\textwidth, height=6.5cm]{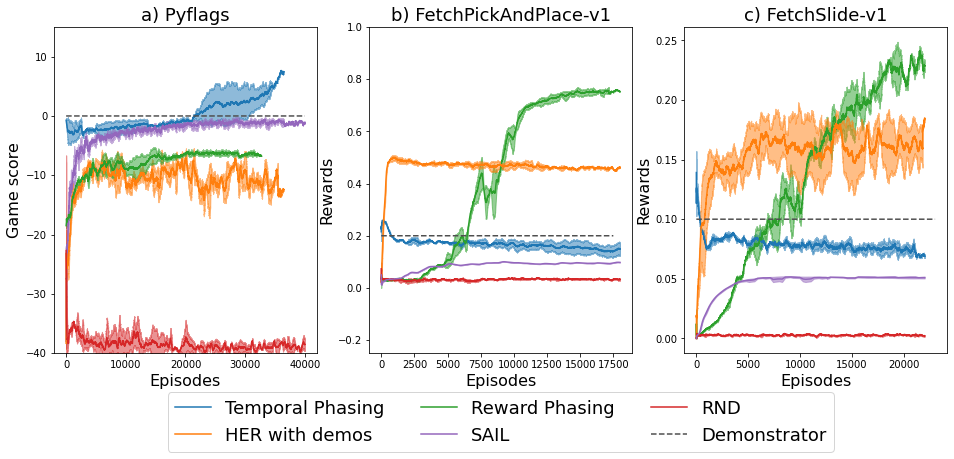}
	\caption{Learning curves of different algorithms tested in the PyFlag, FetchPickAndPlace, and FetchSlide domains. The y-axis represents the average episode score/reward over 300 runs. In the PyFlags domain, the game score is the difference between the number of times the blue and red flags are stolen. In PyFlags 1 episode = 20480 steps; P\&P, FS 1 episode = 1024 steps.
	}
	\label{fig:learning_curves}
\end{figure*}

We start by comparing our task phasing variants against the baseline algorithms. Figure~\ref{fig:learning_curves} presents learning curves for both our best \TP\ (V1: random step) and \RP\ (V1: constant phasing) variants.

\subsubsection{Temporal Phasing}

In the PyFlags domain, see Figure~\ref{fig:learning_curves}(a), \TP\ outperforms all other algorithms with respect to the final performance. \TP\ is also the only approach that is able to outperform the demonstrator.\footnote{In the PyFlags domain, matching the demonstrator's performance is achieved when the average game score is zero (as it is a zero-sum game).} The peak in performance is achieved when the RL policy has full control ($\beta=1$) learning a policy that outperforms the demonstrator (game score $>$ 0). These results provide a positive answer to the experiments' Questions~\ref{Q1},~\ref{Q2}, and~\ref{Q4}.
The results in the Fetch domains, shown in Figure~\ref{fig:learning_curves}(b,c), indicate that \TP\ performed poorly in these domain, with the demonstrator policy only being phased out 50\% on average (not able to achieve the phasing threshold beyond 50\%). This highlights the limitations of the \TP\ approach in learning to perform sensitive tasks that suffer from unrecoverable actions. This is apparent in the P\&P domain where the robotic arm can take one wrong action, such as opening the gripper, leading it to drop the block. A similar phenomenon can be observed in the FS domain where pushing the block in certain directions leads to the block becoming unreachable. In the PyFlags domain, by contrast, 
the demonstrator's policy is more likely to recover from a limited number of bad actions, making the \TP\ approach more appropriate.
These results provide a positive answer to the experiments' Question~\ref{Q3}.

\subsubsection{Reward Phasing}


Figure~\ref{fig:learning_curves} paints a picture where \RP\ achieves state-of-the-art performance in both Fetch domains.
It can be observed that during the phasing process \RP\ results in mostly monotonic improvement (in expectancy), supporting the claims made in Theorem~\ref{theorem:theory_rew_phasing}. 

In the Fetch domains, the policy learned using \RP\ not only learns to perform the task but also outperforms the demonstrator policy. These results provide further positive support to experiments' Questions~\ref{Q1},~\ref{Q2}, and~\ref{Q4}. In the PyFlags domain, although the policy learnt using \RP\ does not outperform the demonstrator policy, we observe that the learned policy is still able to steal the enemy's flag 5 times per episode on average. This indicates that \RP\ was successful in learning a policy that retained useful behaviors in sparse reward settings. However, the RL policy seemed to get stuck in a local minima where it is unable to outperform the opponent. 
This provides further positive support to the experiments' Question~\ref{Q3}.

\subsubsection{HER with demonstrations}

Despite being able to learn a reasonable sub-optimal policy, the HER algorithm did not outperform both our best task phasing approaches in the reported domains.
The results highlight the drawbacks of the HER algorithm in adversarial settings such as the PyFlags domain. The HER algorithm provides a reward to the agent for achieving sub-goals based on how close they are to the true goal state, but it does not consider how good those sub-goals are with respect to the true reward. For instance, the agent may receive a large reward when it reaches a state close to the opponent's flag, but it is not penalized if this state leads the agent directly into the opponent's line of fire.

\subsubsection{Random Network Distillation}

The RND algorithm was not able to perform meaningful learning in any of the reported domains.
This indicates that although exploration-based approaches are known to provide good results, they require a substantial training period compared to algorithms that utilize demonstrations or some form of domain knowledge. Similar trends for RND have been previously observed and reported~\cite{yang2021exploration}. 

\paragraph{Self-adaptive Imitation Learning} 

SAIL was unable to outperform at least one of our task-phasing approaches in all of the reported domains. While in the PyFlags domain, it was able to outperform \RP, it still underperformed \TP.
SAIL exhibited similar limitations to \TP\ in the Fetch domains where it failed to even match the demonstrator's performance. We speculate that similar to \TP, SAIL is unable to effectively deal with unrecoverable actions which prevents it from experiencing sparse (hard to reach) rewards. 


\section{Limitations and Future Work}   \label{Sec:lim_future}

A limitation of task phasing is the need for human input to choose between its two (and potentially more) variants, \TP\ or \RP. Future work aims to improve \RP\ by fusing it with the \textit{Dataset Aggregation}~\cite{ross2011reduction} approach for no-regret online learning. We expect such an approach to outperform \TP\ in all of the domains, making \RP\ a default variant choice. In doing so, however, the availability of an online interactive demonstrator must be assumed. Such an assumption can be motivated by RL domains that target optimizing existing, suboptimal controllers. Examples include Traffic Signal Controllers~\cite{Ault2021signals,sharon2021transportation,Ault20} and Autonomous Driving~\cite{kiran2021deep}.


\section{Summary}

This paper presents two general, domain-independent approaches for designing a curriculum continuum from demonstrations: \TP\ and \RP. The curriculum continuum allows decomposing a complex RL task into a set of tasks with progressively increasing complexity. We show that under the assumption of a continuous (task, optimal policy) space, a task phasing approach with a sufficiently small step size, is guaranteed to learn the optimal policy for any task. We show that the \RP\ curriculum must result in policies that are monotonically non-decreasing with respect to the expected return in the target task. Experimental results in sparse reward domains indicate that \TP\ and/or \RP\ can significantly surpass state-of-the-art algorithms in terms of asymptotic performance.
The results indicate that \TP\ is more applicable for tasks that do not require high precision or are prone to catastrophic actions, and that \RP\ produces sub-optimal results when IRL fails to accurately mimic the demonstrator. 

\section*{Acknowledgements}

\begin{enumerate}
    \item The reported work has taken place in the \textit{PiStar} AI and Optimization Lab at Texas A\&M.  PiStar research is supported in part by NSF (IIS-2238979).

    \item A portion of this work was funded by \textit{Ultra Electronics Advanced Tactical Systems}, Inc. and \textit{Ultra Labs}.

    \item A portion of this work has taken place in the Learning Agents Research Group (LARG) at UT Austin.  LARG research is supported in part by NSF (CPS-1739964, IIS-1724157, FAIN-2019844), ONR (N00014-18-2243), ARO (W911NF-19-2-0333), DARPA, GM, Bosch, and UT Austin's Good Systems grand challenge.  
    
    \item Peter Stone serves as the Executive Director of Sony AI America and receives financial compensation for this work.  The terms of this arrangement have been reviewed and approved by the University of Texas at Austin in accordance with its policy on objectivity in research.

    \item A portion of this work was funded by the Mays Innovation Research Center at Texas A\&M
\end{enumerate}

\bibliography{aaai23}
\appendix
\counterwithin{figure}{section}
\counterwithin{table}{section}
 
\section{Appendix}

\subsection{Domain Description} \label{Sec:domain}
We demonstrate the effectiveness of the proposed algorithms in the following domains:

\begin{figure*}[ht!]
    \centering
    \begin{subfigure}{0.28\textwidth}
    \centering
	\includegraphics[width=\textwidth, height=\textwidth]{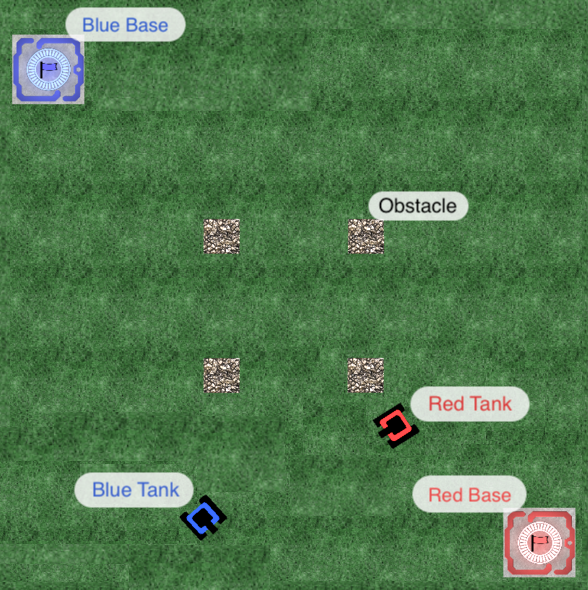}
	\label{fig:CTF env}
	\end{subfigure}%
	\begin{subfigure}{0.28\textwidth}
    \centering
	\includegraphics[width=\textwidth, height=\textwidth]{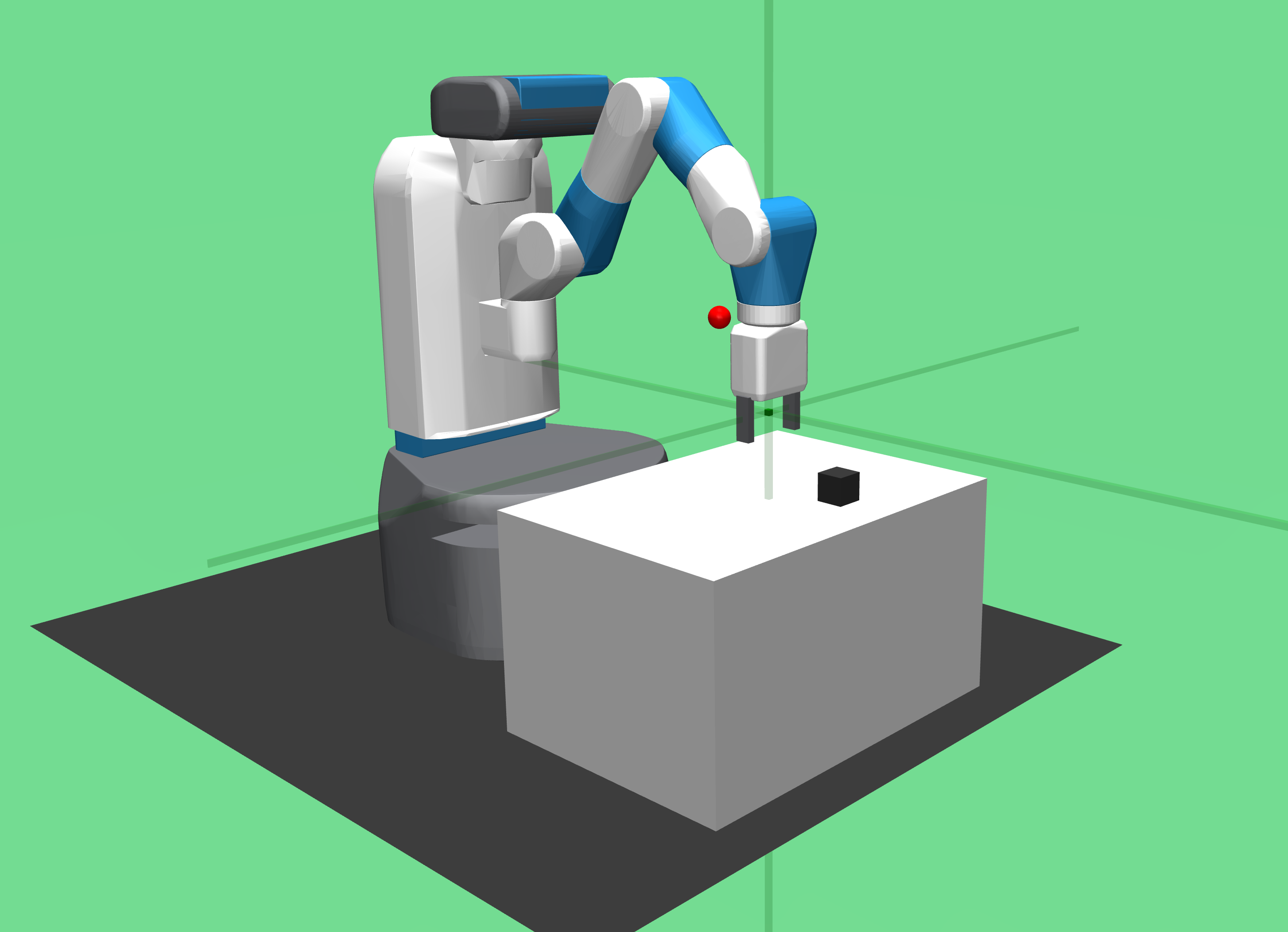}
	\label{fig:Fetch}
	\end{subfigure}%
	\begin{subfigure}{0.28\textwidth}
    \centering
	\includegraphics[width=\textwidth, height=\textwidth]{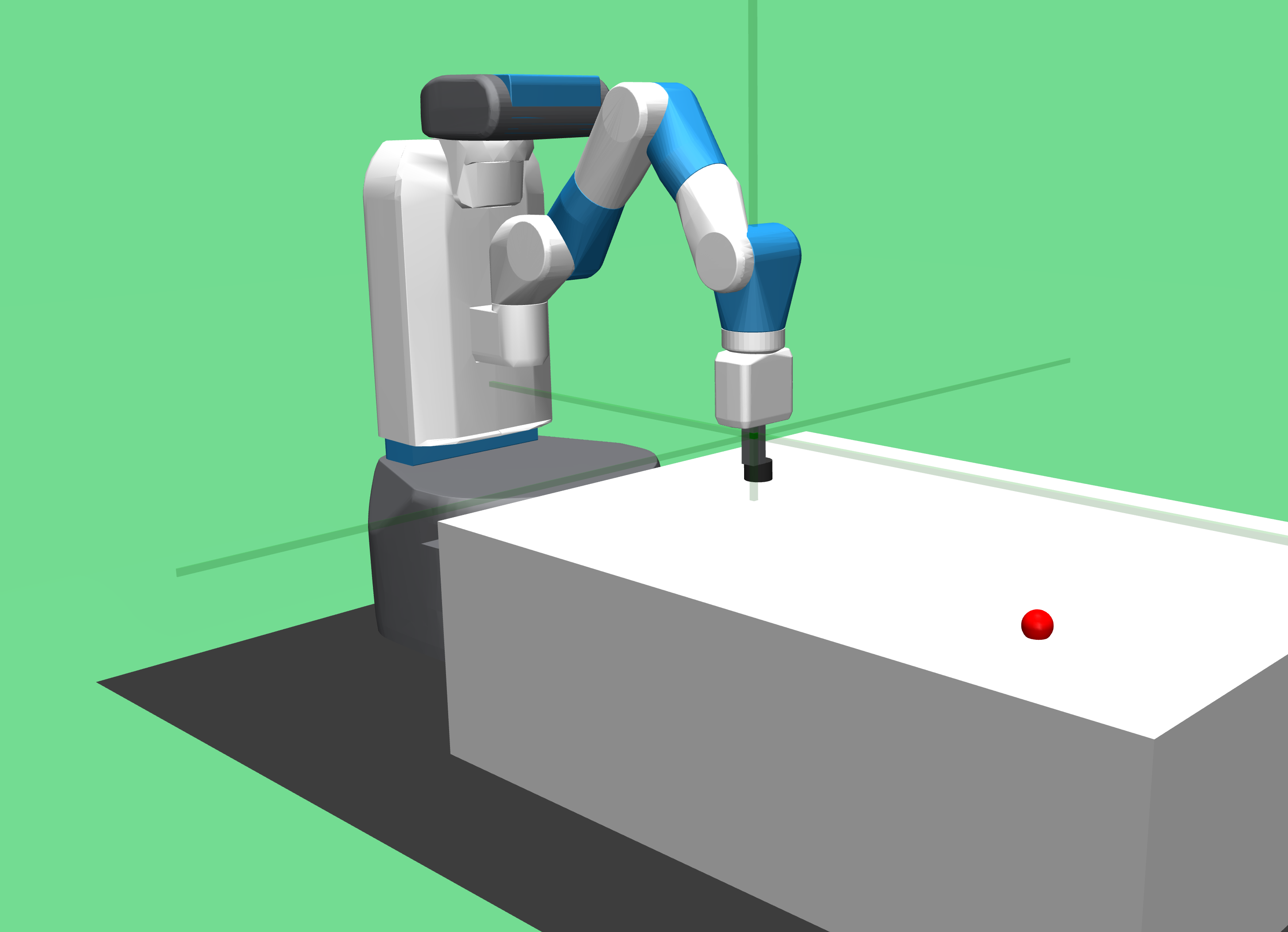}
	\label{fig:FetchSlide}
	\end{subfigure}%
	\caption{(a) The PyFlags domain; (b) FetchPickAndPlace-v1 domain; (c) FetchSlide-v1 domain}
	\label{Fig:Domains}
\end{figure*}

\subsubsection{PyFlags}

PyFlags~\cite{pyflags} is a domain consisting of two adversarial tanks playing a game of capture the flag. This domain was chosen as it provides a standard rule-based player, which can be used as a sub-optimal demonstrator. The domain provides a zero-sum, adversarial game that allows for an intuitive policy comparison (positive score = RL-controlled agent is better than the opponent, and vice versa). This domain also exhibits a continuous control task with sparse rewards, making it challenging for an RL agent to learn an optimized policy. Effective exploration in this domain is especially challenging given that the opponent is actively targeting the RL-controlled agent. Reaching the opponent's base without being shot, stealing the flag, and returning it to the home base is highly unlikely following random exploration.

The PyFlags domain is illustrated in Figure~\ref{Fig:Domains}(a). The RL agent controls the blue tank, whereas the red tank is the opponent (a rule-based player provided by the domain). The domain is defined by:

\begin{itemize}
    \item \textbf{State space, $S$.}
The positions and orientations of all the objects in the environment (obstacles, bases, flags, and red tank), relative to the blue tank. The state space also includes the angular and linear velocity of the blue tank and a time-to-reload value for both tanks, representing the earliest time (in the future) at which the tank can shoot. The last 4 observation frames are provided at each time step.

    \item \textbf{Action space, $\mathcal{A}$.}
The action space is a Cartesian product of 3 values: angular velocity $\in [-1,1]$, linear velocity $\in [0,1]$, and fire $\in \{0,1\}$. These actions affect the relevant state values for the blue tank (the tank controlled by the RL agent). These actions are available on time steps where the RL-agent is assigned control.

    \item \textbf{Transition function, $\mathcal{P}$.}
States evolve based on actions assigned to the blue tank and based on actions assigned to the red tank. The red tank actions are assumed to be drawn from an unknown adversarial policy. If either tank gets shot it re-spawns at its own base after a randomly selected re-spawn period, between 100 and 200 time steps.

    \item \textbf{Target reward function, $R^f$.}
A positive reward (+1) is received when the blue tank returns to its base with the red flag, and a negative reward (-1) is received when the red tank does the same.

\end{itemize}

\subsubsection{FetchPickAndPlace-v1} 

FetchPickAndPlace~\cite{Plappert2018MultiGoalRL}, is a robotics domain consisting of a 7-DoF Fetch robotics arm, which has a two-fingered parallel gripper. The task in this domain is to use the gripper to hold a box and carry it to the target position, which may be on the table or in the air above it. It is a multi-goal domain since the target position is chosen randomly in every episode. This domain was chosen as it is well suited to demonstrate the algorithm's capability of solving continuous control, hard-exploration sparse reward tasks. The FetchPickAndPlace environment is the most complex of the Fetch environments, as it is difficult to learn to grip the object using random actions. A standard rule-based player ~\cite{nair2018overcoming} is also provided, which can be used as a suboptimal demonstrator.

The FetchPickAndPlace domain is illustrated in Figure~\ref{Fig:Domains}(b). The domain is defined by:

\begin{itemize}
    \item \textbf{State space, $S$.}
Observations include the Cartesian position of the gripper, its
linear velocity as well as the position and linear velocity of the robot’s gripper. It also includes the object’s Cartesian position and rotation using Euler angles, its linear and angular
velocities, its position and linear velocities relative to the gripper, as well as the desired target coordinates it must be moved to. The goals that are achieved by the agent so far can also be obtained, which is required by the HER algorithm.

    \item \textbf{Action space, $\mathcal{A}$.}
The action space is 4-dimensional: 3 dimensions specify the desired gripper movement in Cartesian coordinates $\in [-1,1]$ and
the last dimension controls opening and closing of the gripper $\in [-1,1]$.

    \item \textbf{Transition function, $\mathcal{P}$.}
The object and its desired target coordinates are randomly selected at the beginning of every episode. The robotic arm and its gripper change their position after each step, based on the action taken. The same action is applied for 20 subsequent simulator steps (with $\Delta t = 0.002$ each) before returning control to the agent.

    \item \textbf{Target reward function, $R^f$.}
The rewards are sparse and binary. A positive reward (+1) is received when the object is within 5cm of the target coordinates and 0 otherwise.

\end{itemize}

\subsubsection{FetchSlide-v1} 

FetchSlide~\cite{Plappert2018MultiGoalRL}, is a robotics domain consisting of a 7-DoF Fetch robotics arm, which has a two-fingered end effector used for pushing a block on a long table top. The reach of the robotic arm is limited to only a small portion of the table. The task in this domain is to use the end effector of the robotic arm to push the box in a precise manner, such that it comes to a stop at a target location on the table. The target location may not be within reach of the robotic arm, requiring it to push the block with an exact force and angle such that it slides exactly to the target location. It is a multi-goal domain since the target position is chosen randomly in every episode. This domain was chosen as it is well suited to demonstrate the algorithm's capability of solving continuous control, hard-exploration sparse reward tasks. The FetchSlide environment is a more complex version of the FetchPush environment, as it requires greater precision. A standard rule-based player ~\cite{nair2018overcoming} is also provided, which can be used as a suboptimal demonstrator.

The FetchSlide domain is illustrated in Figure~\ref{Fig:Domains}(c). The domain is defined by:

\begin{itemize}
    \item \textbf{State space, $S$.}
Observations include the Cartesian position of the gripper, its
linear velocity as well as the position and linear velocity of the robot’s gripper. It also includes the object’s Cartesian position and rotation using Euler angles, its linear and angular
velocities, its position and linear velocities relative to the gripper, as well as the desired target coordinates it must be moved to. The goals that are achieved by the agent so far can also be obtained, which is required by the HER algorithm.

    \item \textbf{Action space, $\mathcal{A}$.}
The action space is 4-dimensional: 3 dimensions specify the desired gripper movement in Cartesian coordinates $\in [-1,1]$ and
the last dimension controls opening and closing of the gripper $\in [-1,1]$. In this domain, the gripper action is disabled.

    \item \textbf{Transition function, $\mathcal{P}$.}
The object and its target coordinates are randomly selected at the beginning of every episode. The robotic arm and its gripper change their position after each step, based on the action taken. The same action is applied for 20 subsequent simulator steps (with $\Delta t = 0.002$) before returning control to the agent.

    \item \textbf{Target reward function, $R^f$.}
The rewards are sparse and binary. A positive reward (+1) is received when the object is within 5cm of the target coordinates and 0 otherwise.

\end{itemize}

\subsection{Hyperparameters}   \label{Sec:Hyperparams}

The hyperparameters of the algorithms used in the experiments in the PyFlags, FetchPickAndPlace and FetchSlide domains are mentioned below. The network architecture specifies the number of hidden layers in the neural network and the number of nodes each layer is composed of. The $\tanh$ activation function is applied on each node of the hidden layers. They are shared by the actor and critic networks. The learning rate mentioned applies to both the actor and critic networks. The demonstrations for all the algorithms requiring demonstrations are collected from a rule-based policy for each domain.

\begin{table*}[ht!]
    \centering
    \begin{tabular}{|c||c|c|c|c|c|c|c|c|c|}
        \hline
            &  Network arch. & $\gamma$ & $\lambda$ & lr & $clipping(\epsilon)$ & ent\_coef & batch\_size & $\alpha$\\ [0.5ex] 
        \hline
        PyFlags &  (512,512) & 0.99 & 0.94 & 0.00003 & 0.3 & 0.005 & 20480 & 0.1\\ [0.5ex] 
        \hline 
        P\&P \& FS & (256,256,256) & 0.98 & 0.95 & 0.0003 & 0.2 & 0.005 & 1024 & 0.015\\ [0.5ex] 
        \hline
    \end{tabular}
    \caption{Temporal phasing: PPO hyperparameters for each domain}
    \label{tab:Temporal_Phasing}
\end{table*}

\begin{table*}[ht!]
    \centering
    \begin{tabular}{|c||c|c|c|c|c|c|c|c|c|}
        \hline
            &  Network arch. & $\gamma$ & $\lambda$ & lr & $clipping(\epsilon)$ & ent\_coef & batch\_size & $\alpha$ & No. Demos\\ [0.5ex] 
        \hline
        PyFlags &  (512,512) & 0.99 & 0.95 & 0.0003 & 0.2 & 0.005 & 20480 & 0.05 & 50\\ [0.5ex] 
        \hline 
        P\&P \& FS & (256,256,256) & 0.98 & 0.95 & 0.0003 & 0.2 & 0.005 & 1024 & 0.015 & 50\\ [0.5ex] 
        \hline
    \end{tabular}
    \caption{Reward Phasing: PPO hyperparameters for each domain}
    \label{tab:Reward_Phasing}
\end{table*}

\begin{table*}[ht!]
    \centering
    \begin{tabular}{|c||c|c|c|c|c|c|c|c|c|}
        \hline
            &  Network arch. & $\gamma$ & lr & batch\_size & activation func.\\ [0.5ex] 
        \hline
        PyFlags &  (128,128) & 0.99 & 0.0003 & 20480 & ReLU\\ [0.5ex] 
        \hline 
        P\&P \& FS & (128,128) & 0.98 & 0.0003 & 1024 & ReLU\\ [0.5ex] 
        \hline
    \end{tabular}
    \caption{Reward Phasing: IRL discriminator hyperparameters for each domain}
    \label{tab:IRL}
\end{table*}

\begin{table*}[ht!]
    \centering
    \begin{tabular}{|c||c|c|c|c|}
        \hline
            &  Network arch. & Q\_lr & pi\_lr & No. Demos\\ [0.5ex] 
        \hline
        PyFlags &  (512,512) & 0.0001 & 0.0001 & 50\\ [0.5ex] 
        \hline 
    \end{tabular}
    \caption{HER with Demos: Modified HER hyperparameters for PyFlags domain}
    \label{tab:HER}
\end{table*}

\begin{table*}[h!]
    \centering
    \begin{tabular}{|c||c|c|c|c|c|c|c|}
        \hline
            &  Network arch. & $\gamma$ & $\lambda$ & lr & $clipping(\epsilon)$ & ent\_coef & batch\_size\\ [0.5ex]
        \hline
        PyFlags &  (512,512) & 0.99 & 0.95 & 0.0003 & 0.2 & 0.005 & 20480\\ [0.5ex] 
        \hline 
        P\&P \& FS & (256,256,256) & 0.98 & 0.95 & 0.0003 & 0.2 & 0.005 & 1024\\ [0.5ex] 
        \hline
    \end{tabular}
    \caption{RND: PPO hyperparameters for each domain}
    \label{tab:RND}
\end{table*}

\begin{table*}[h!]
    \centering
    \begin{tabular}{|c||c|c|c|c|c|c|c|}
        \hline
            &  Network arch. & $\gamma$ & buffer size & Demo buffer size & No.Demos & batch\_size & train freq.\\ [0.5ex]
        \hline
        PyFlags &  (512,512) & 0.99 & 1000000 & 600000 & 50 & 2048 & 20480\\ [0.5ex] 
        \hline 
        P\&P \& FS & (256,256,256) & 0.98 & 1000000 & 50000 & 50 & 256 & 1024\\ [0.5ex] 
        \hline
    \end{tabular}
    \caption{SAIL: TD3 hyperparameters for each domain}
    \label{tab:SAIL}
\end{table*}

\subsubsection{Temporal Phasing} \label{Sec:Temporal_hyperparams}

In the PyFlags domain we set $\alpha=0.1$ if the current RL policy achieved a game score $\geq$ 0 in the last 50 episodes; else we set $\alpha=0$. Since the demonstrator used is the same as the opponents policy a threshold of 0 implies that the performance must be at least as good as that of the demonstrator. In the P\&P domain we set $\alpha=0.015$ if the current RL policy achieved a game score $\geq$ 0.2 in the last 50 episodes; else we set $\alpha=0$. A 0.2 threshold would imply that the agent was able to reach the goal position on average. Similarly, $\alpha=0.015$ in the FS domain if the game score $\geq$ 0.1 in the last 50 episodes; else we set $\alpha=0$.

\subsubsection{Reward Phasing}

In the P\&P and FS domains, we find that reducing the entropy coefficient to 0.001, the learning rate to 0.00007 and $\alpha$ to 0.001 after 75\% of reward phasing results in more stable learning. This may be interpreted as a need to prevent the policy from drifting too far from the current policy as the demonstrators affect on the policy diminishes. A large drift may cause the policy to forget useful behaviours that it learnt while imitating the demonstrator.

\subsubsection{Hindsight Experience Replay (HER) with demonstrations}

The hyperparameters for FetchPickAndPlace and FetchSlide are the same as those mentioned in ~\cite{nair2018overcoming}. Similarly, most of the hyperparameters for PyFlags are the same as those mentioned in ~\cite{nair2018overcoming} with a few adjustments so that the algorithm better adapts to the domain. Each demonstration in the PyFlags domain consists of 20480 timesteps. HER with demonstrations baseline uses DDPG~\cite{Lillicrap2016ContinuousCW} as the policy optimization algorithm.

\subsubsection{Random Network Distillation (RND)~\cite{burda2018exploration}.}

Proximal Policy Optimization (PPO)~\cite{PPO} is used as the policy optimization algorithm. The hidden layers of the neural network are shared by an actor, intrinsic value function, and extrinsic value function. The feature predictor neural network as well as the fixed randomly initialized neural network consists of 2 hidden layers, each composed of 32 nodes with $\tanh$ activation function, and an output feature layer consisting of 16 nodes, in all domains.

\subsubsection{Self-Adaptive Imitation Learning (SAIL)~\cite{zhu2020learning}.}

Twin-delayed deep deterministic policy gradient (TD3)~\cite{fujimoto2018addressing} is used as the policy optimizing algorithm, The hyper-parameters are tuned using the default optimizer tool provided in the reproducible code, presented in ~\cite{zhu2020learning}. The network architecture and demo buffer information is provided in Table ~\ref{tab:SAIL}.

\subsection{Phasing Variants} \label{Sec:variants}


\subsubsection{\TP\ variants}

\begin{figure}[]
    \centering
    \includegraphics[width=0.5\textwidth]{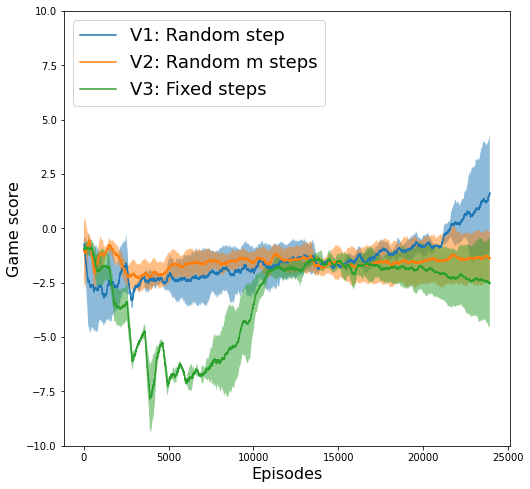}
    \caption{Learning trend comparisons between the task phasing variants in PyFlags. Curves are averaged over 300 runs.}
	\label{fig:temporal_variants}
\end{figure}

Since \TP\ was only successful in the PyFlags domain, we present the comparison of its different variants and the ablation studies in this domain. 
The learning trends of the \TP\ variants can be observed in Figure~\ref{fig:temporal_variants}. Variant 1 is able to complete \TP\ (reach $\beta=1$) and surpass the demonstrator's policy, whereas variants 2 and 3 did not reach $\beta=1$ following 30,000 episodes. Recall that a dynamic step size was used for these variants hence reaching $\beta=1$ is not guaranteed. 

\begin{figure}[]
    \centering
    \includegraphics[width=0.5\textwidth]{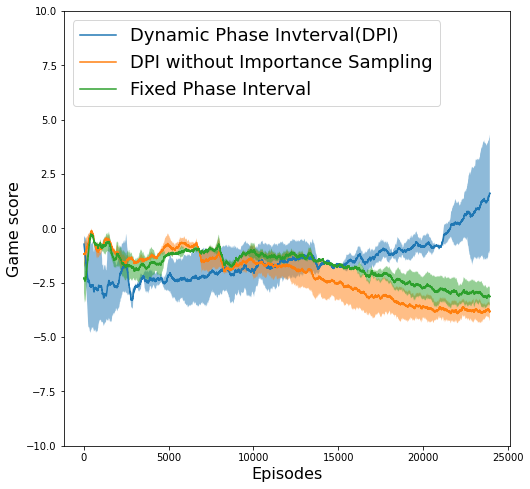}
    \caption{Temporal phasing with/without dynamic step size and importance sampling in PyFlags}
	\label{fig:temporal_mods}
\end{figure}

In Fig.~\ref{fig:temporal_mods} we provide results for the ablation studies conducted on the temporal phasing approach, which demonstrates that Importance Sampling and a dynamic phase interval play a crucial role in the success of this approach. In the fixed-step phasing approach, the sampling frequency of the RL policy is increased by 10\% every 1,000 episodes. Despite providing a long interval between phasing steps, we see that the policy does not perform very well. This is because the policy may often have a long and unstable re-training phase before more control can be given to the RL policy, as can be observed when $\beta=[0.5,0.7]$ (50-70\% RL control) for a dynamic step-size. We attribute this to the fact that the \{$\mathcal{K}_{\beta},\pi^*_\beta$\} space is not continuous. The RL agent is capable of learning to perform the task if a long enough interval is provided to the `re-train' function, (Line~\ref{ln:retrain}, in Algorithm~\ref{alg:phasing}).

\subsubsection{\RP\ variants}

\begin{figure}[]
    \centering
    \includegraphics[width=0.5\textwidth]{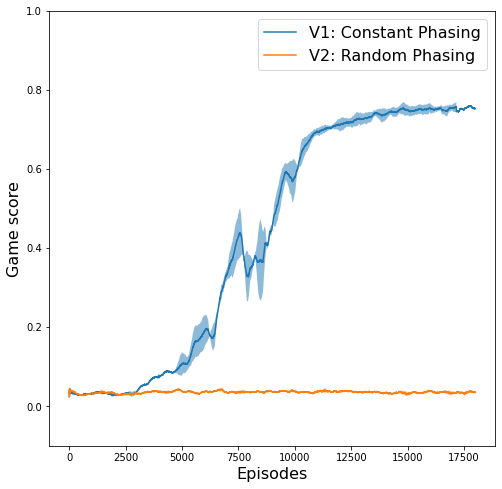}
    \caption{Reward phasing variants in P\&P domain}
	\label{fig:reward_variants}
\end{figure}

\begin{figure}[]
    \centering
    \includegraphics[width=0.5\textwidth]{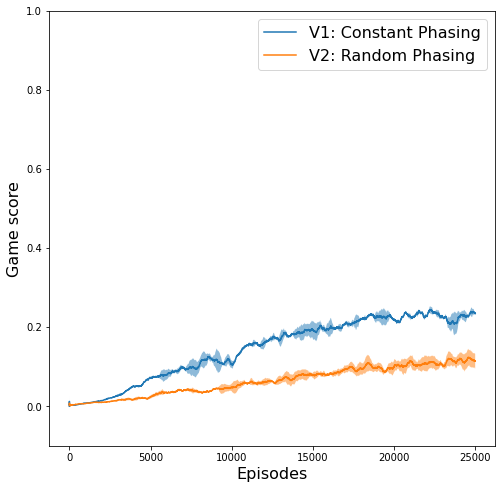}
    \caption{Reward phasing variants in FetchSlide domain}
	\label{fig:reward_variants_fs}
\end{figure}

Since \RP\ was only successful in the FetchPickAndPlace domain, we present the comparison of its different variants and the ablation studies in this domain. The results in Figure~\ref{fig:reward_variants} and Figure~\ref{fig:reward_variants_fs} suggest that V1 (constant phasing) outperforms V2 (random phasing). This indicates that \RP\ works well when the magnitude of the dense reward is gradually reduced instead of reducing the frequency of the dense reward.

\end{document}